\documentclass[10pt, onecolumn, conference]{IEEEtran}

\usepackage{graphicx,url,tabularx,booktabs}
\usepackage[utf8]{inputenc}  
\usepackage{xspace}

\title{Reducing Instability in Synthetic Data Evaluation with a Super-Metric in MalDataGen}

\author{
\IEEEauthorblockN{
    Anna Luiza Gomes da Silva\IEEEauthorrefmark{1},
    Diego Kreutz\IEEEauthorrefmark{1},
    Angelo Diniz\IEEEauthorrefmark{1},\\
    Rodrigo Mansilha\IEEEauthorrefmark{1},
    Celso Nobre da Fonseca\IEEEauthorrefmark{1}
}
\IEEEauthorblockA{\IEEEauthorrefmark{1}Horizon IA Labs and PPGES -- Federal University of Pampa (UNIPAMPA)}
}

\usepackage{minted}

\begin{document} 

\maketitle

\begin{abstract}
    
Evaluating the quality of synthetic data remains a persistent challenge in the Android malware domain due to instability and the lack of standardization among existing metrics. This work integrates into MalDataGen a Super-Metric that aggregates eight metrics across four fidelity dimensions, producing a single weighted score. Experiments involving ten generative models and five balanced datasets demonstrate that the Super-Metric is more stable and consistent than traditional metrics, exhibiting stronger correlations with the actual performance of classifiers.
\end{abstract} 


\section{Introduction}

Synthetic data generation has become an increasingly relevant strategy in cybersecurity \cite{figueira2022survey,lee2025synthetic,hao2024synthetic}, particularly as a way to mitigate the scarcity of real, complete, and high-quality datasets that limit the performance and generalization of machine learning models. Despite these advances, assessing the quality of synthetic data remains a complex and largely non-standardized methodological challenge \cite{HoldoutFrontiers2021}, with no clear consensus on which metrics should be used or how to combine them consistently.

The literature reports a significant fragmentation in the application of fidelity metrics, with studies identifying more than 65 distinct indicators used independently to assess fidelity \cite{super_metrica_sbseg}. This diversity hinders model-to-model comparison, reduces experimental reproducibility, and complicates the integrated interpretation of data quality. Tools such as the Synthetic Data Vault (SDV)\footnote{\url{https://sdv.dev}}, which implements Copula, TVAE, and CTGAN \cite{SDV}; YData Synthetic\footnote{\url{https://ydata.ai/}}, which offers multiple variations of GANs; and Gretel Synthetics\footnote{\url{https://gretel.ai/}}, which uses models such as DGAN, DPGAN, and ACTGAN, attempt to consolidate generation and evaluation processes. Additionally, initiatives such as \cite{SynSys} demonstrate the application of HMMs for time-series generation in the healthcare domain. However, these platforms exhibit limitations related to flexibility, restricted customization capabilities, and relatively small sets of pre-implemented algorithms.

To overcome these limitations, this work enhances the MalDataGen framework \cite{maldatagen}, a modular open-source platform designed for the generation of synthetic tabular data, through the integration of a generalizable super-metric developed to unify fidelity evaluation \cite{super_metrica_sbseg}. The super-metric aggregates eight metrics distributed across four fundamental dimensions—Distance, Correlation/Association, Feature Similarity, and Multivariate Distribution—producing a single weighted score that reduces the variability and inconsistency observed in evaluations based on isolated metrics.

The central contribution of this work is transforming MalDataGen from a generation tool into a complete ecosystem for multidimensional generation and evaluation of synthetic data aimed at Android malware detection. By incorporating the super-metric, the framework provides a more robust, consistent, and contextualized evaluation, making it more suitable for critical cybersecurity applications.

\section{The MalDataGen Framework}

MalDataGen is a modular and open-source framework designed to systematically and reproducibly orchestrate the generation and evaluation of synthetic tabular data in the context of Android malware detection. Its goal is to provide a unified platform that enables the comparison of different generative models under the same experimental methodology, reducing implementation bias and ensuring consistency across executions.

The framework’s architecture is organized into three main components:: (i) \textit{input and preprocessing}, esponsible for standardizing \emph{datasets}, normalizing attributes, and preparing data for the generative models;; (ii) \textit{generative layer}, which integrates multiple families of models capable of synthesizing tabular datasets with varying levels of complexity; and (iii) \textit{evaluation layer}, which computes traditional fidelity and utility metrics, as well as the Super-Metric integrated in this work.

A central pillar of MalDataGen is its function as a flexible benchmark, enabling different generation paradigms to be evaluated under identical conditions. To support this, the generative layer includes four groups of models:
(i) \textit{Adversarial Models} (GANs): classical GAN, WGAN, and WGAN-GP;  
(ii) \textit{Autoencoders}: standard autoencoder, VAE, and quantized VAE;  
(iii) \textit{Diffusion Models}: Denoising Diffusion and Latent Diffusion;  
(iv) \textit{Statistical and third-party models}: SMOTE and SDV library models (CTGAN, TVAE, Copula).

\begin{figure}[htb!]
    \centering
    \includegraphics[width=0.90\linewidth]{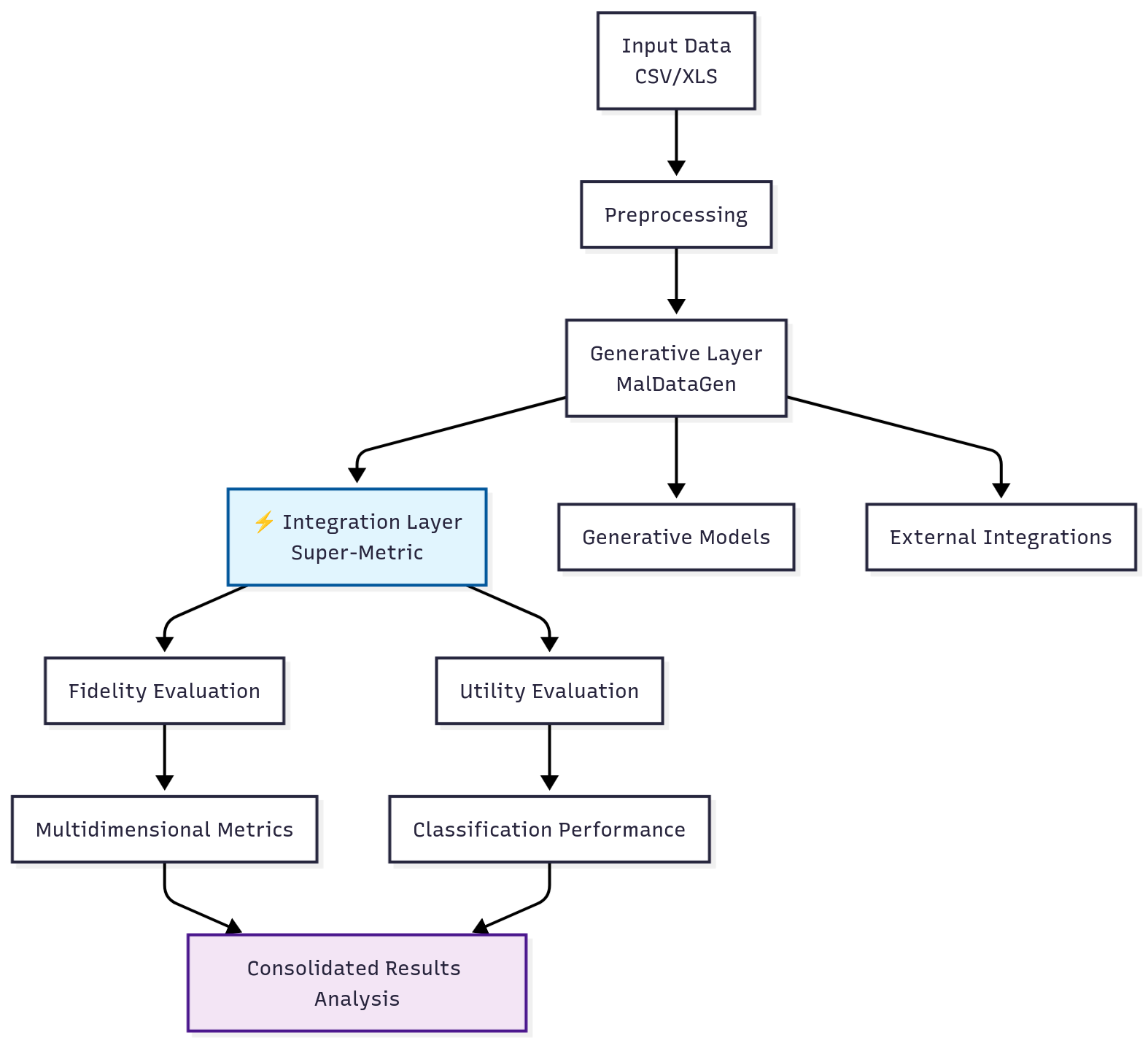}
    \caption{Workflow of the synthetic data generation and evaluation methodology in MalDataGen.}
    \label{fig:fluxograma}
\end{figure}

Figure~\ref{fig:fluxograma} presents a diagram illustrating the workflow across these components.

The metrics module in MalDataGen organizes the evaluation process into three categories: binary metrics (e.g., precision, recall, F1-score), distance metrics (such as Euclidean and Hellinger), and probabilistic metrics (such as AUC-ROC). The internal infrastructure standardizes result storage by evaluation strategy, classifier, and fold, ensuring traceability and comparability across experiments.

The Super-Metric, integrated as a composite metric within the distance module, extends the evaluation system by providing a consolidated multidimensional analysis. It combines eight metrics distributed across four fundamental dimensions—distance, association, feature similarity, and multivariate distribution—and produces a single weighted final score. Its integration occurs transparently within the framework’s internal workflow, using the same routines and data structures as conventional metrics.

With this integration, MalDataGen evolves from a data generation tool into a complete ecosystem for generating, evaluating, and rigorously comparing generative models applied to the Android malware domain. This enables more consistent, stable, and comparable analyses, contributing to reproducible and methodologically sound experiments in cybersecurity.

\section{The Super-Metric}

Evaluating the quality of synthetic data remains a central challenge in the generation of tabular datasets, as traditional fidelity and utility metrics—such as statistical distances, distribution divergences, and association measures—tend to capture only specific aspects of the problem. These metrics are generally domain-sensitive, exhibit instability in the presence of multimodal distributions, and can be difficult to interpret collectively, especially in scenarios characterized by strong class imbalance and highly sparse binary attributes, as commonly observed in Android malware data. Although composite approaches such as TabSynDex \cite{tabsyndex} represent advances by consolidating multiple dimensions into a single index, previous studies \cite{super_metrica_sbseg} indicate that they may still present significant variation across generative models and do not always reflect the real impact of synthetic data on the performance of supervised classifiers.

In this context, the Super-Metric was designed to provide a more robust and informative alternative. Its formulation combines different fidelity dimensions in a weighted manner and automatically adjusts the weights to maximize correlation with utility metrics such as \emph{recall} and F1-score. Rather than relying on uniform aggregation, it operates as an optimized composition that prioritizes the most relevant dimensions for malware detection tasks, capturing both structural similarity and the preservation of discriminative patterns. In this way, the Super-Metric simultaneously functions as a global indicator of quality and as a predictive estimator of the practical effectiveness of synthetic data in real classification scenarios.

\section{Evaluation}

To assess the effectiveness of the Super-Metric integrated into MalDataGen, we conducted experiments involving ten generative models and five balanced Android malware \emph{datasets} \cite{nawshin2024malware,alomari2023malware}. For each combination of \emph{dataset} and generator, we computed traditional fidelity metrics as well as the proposed Super-Metric, comparing them against utility metrics (recall and F1-score) obtained from classifiers trained on synthetic data and evaluated exclusively on real data.

In all experiments, the Super-Metric was computed separately for each dataset, with the goal of simultaneously reducing the gap between recall and F1-score values. This approach aims to ensure that the final score reflects not only statistical fidelity but also the practical utility of synthetic data in real classification tasks.

Two main types of visualizations were produced: (i) heatmaps showing the average correlation between each fidelity metric and the utility metrics, and (ii) boxplots representing the distribution of these correlations across different generators. Based on these analyses, we evaluated three desirable properties for a fidelity metric: consistency, understood as maintaining the same correlation sign; stability, related to low variance across models; and robustness, defined as behavior independent of the generative architecture.

\begin{figure}[h!]
    \centering
    \includegraphics[width=1\linewidth]{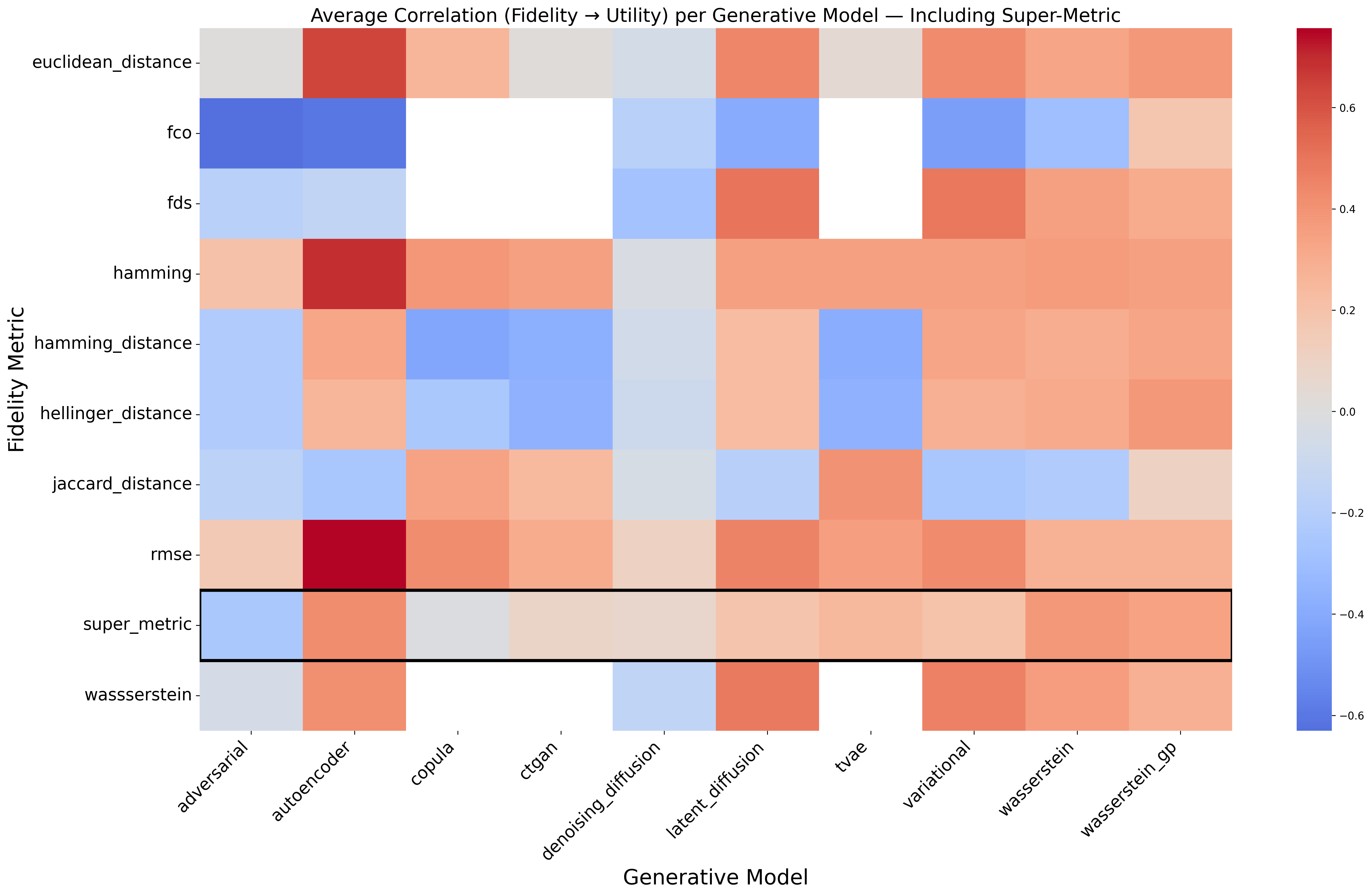}
    \caption{Heatmap – Average correlation between fidelity metrics and utility metrics (recall and F1-score) per generative model.}
    \label{fig:heatmap}
\end{figure}

\begin{figure}[h!]
    \centering
    \includegraphics[width=\linewidth]{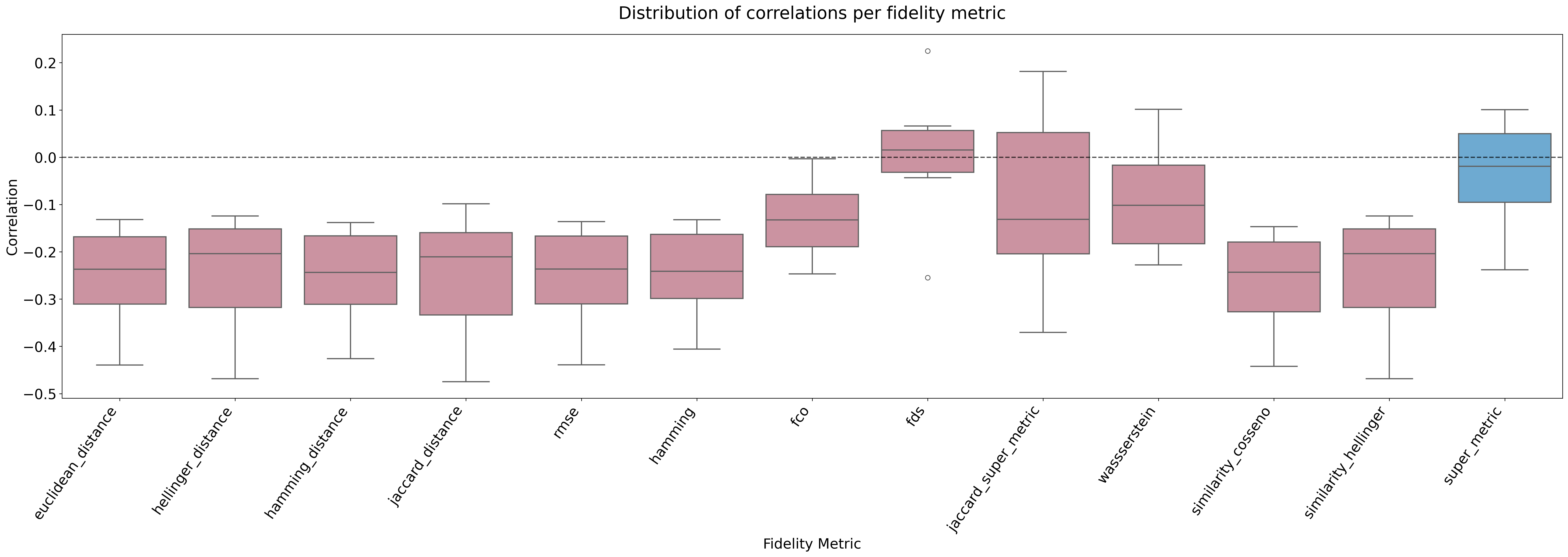}
    \caption{Boxplot – Distribution of the correlation between fidelity metrics and utility metrics (recall and F1-score) per generative model.}
    \label{fig:boxplot}
\end{figure}

Figures \ref{fig:heatmap} and \ref{fig:boxplot} illustrate these results and highlight the advantage of the Super-Metric in heterogeneous generation scenarios. The results show that traditional metrics exhibit highly unstable behavior, alternating between positive and negative correlations and displaying large variation across generative models. This behavior indicates that none of these metrics can serve as a universal fidelity metric. In contrast, the Super-Metric demonstrates greater stability, a consistent correlation sign, and better alignment with recall and F1-score. Even when it does not achieve the highest absolute correlation, it stands out as the most stable metric across generators, indicating that its weighted aggregation reduces noise and mitigates limitations present in metrics evaluated in isolation.

\section{Final Considerations and Future Work}

This work integrated a fidelity-oriented Super-Metric into MalDataGen, extending the framework beyond synthetic data generation and consolidating it as a robust benchmarking platform. The diversity of supported models—including GANs, autoencoders, statistical methods, and diffusion models—enabled a comprehensive comparative analysis across heterogeneous generation scenarios. The experiments showed that traditional fidelity metrics exhibit inconsistent behaviors and high sensitivity to the type of generator, resulting in low stability and fluctuating correlations with utility metrics. In contrast, the Super-Metric demonstrated greater consistency, lower variance across models, and better alignment with the real performance of classifiers, addressing an important gap in the evaluation of synthetic data quality within the malware domain.

For future work, there are several opportunities to enhance the Super-Metric using advanced optimization techniques, such as evolutionary algorithms, meta-learning, and nonlinear strategies for metric combination. Expanding the approach to other domains and introducing interpretability mechanisms also emerge as promising directions, enabling a deeper understanding of the contribution of each dimension of the Super-Metric. Furthermore, its integration into MLOps pipelines creates opportunities for continuous monitoring of synthetic data quality in real production environments, reinforcing the potential of the metric as a central component within the MalDataGen ecosystem.

\bibliographystyle{ieeetr}
\bibliography{references}

\end{document}